\documentclass[runningheads]{llncs}
\usepackage{graphicx}
\input epsf
\begin{document}
\title{Visual-based Positioning and Pose Estimation} 

\author{Somnuk Phon-Amnuaisuk\inst{1,2}\orcidID{0000-0003-2130-185X} \and Ken T. Murata\inst{3} \and La-Or Kovavisaruch\inst{4} \and Tiong-Hoo Lim\inst{1} \and Praphan Pavarangkoon\inst{3} \and Takamichi Mizuhara\inst{5}}

\authorrunning{S. Phon-Amnuaisuk et al.}

\institute{Media Informatics Special Interest Group, Universiti Teknologi Brunei, Brunei \and School of Computing and Information Technology, Universiti Teknologi Brunei \and National Institute of Information and Communications Technology, Tokyo, Japan \and National Electronics and Computer Technology Center (NECTEC), Thailand \and CLEALINKTECHNOLOGY Co., Ltd., Kyoto, Japan \\
\email{somnuk.phonamnuaisuk; lim.tiong.hoo@utb.edu.bn; ken.murata; praphan@nict.go.jp; La-Or.Kovavisaruch@nectec.or.th; mizuhara@clealink.jp}
}

\maketitle

\begin{abstract}
Recent advances in deep learning and computer vision offer an excellent opportunity to investigate high-level visual analysis tasks such as human localization and human pose estimation. Although the performance of human localization and human pose estimation have significantly improved in recent reports, they are not perfect and erroneous localization and pose estimation can be expected among video frames. Studies on the integration of these techniques into a generic pipeline that is robust to noise introduced from those errors are still lacking.  This paper fills the missing study. We explored and developed two working pipelines that suited the visual-based positioning and pose estimation tasks. Analyses of the proposed pipelines were conducted on a badminton game. We showed that the concept of tracking by detection could work well, and errors in position and pose could be effectively handled by a linear interpolation technique using information from nearby frames. The results showed that the \emph{Visual-based Positioning and Pose Estimation} could deliver position and pose estimations with good spatial and temporal resolutions. 

\end{abstract}

\section{Introduction}
Recent progress in object detection and tracking has opened up various potential areas of applications such as visual-based indoor position analyses, visitors' hot spot analyses, and visitors' activity analyses from pose information.
In this paper, a study of a badminton game analyses is discussed. The players' positions on the badminton court and their skeleton poses\footnote{Action analysis based on a skeleton figure, i.e., a stick man figure.} were extracted from a sequence of images. This work proposed a pipeline of deep learning and computer vision tasks that extracted players' movements and actions from an image sequence.
In summary, players' positions were detected and tracked using Mask R-CNN. The positions read from a 2D camera view coordinate were transformed into positions in 2D top-view in the world coordinate. The players' positions across a frame sequence would reveal the players' movement over time; analyzing each player's pose would then reveal their actions. These two sources of information could be employed in further high-level visual analytic tasks.

We coined the following two terms to facilitate the presentation and discussion in this paper: Outside-in Vision-based Positioning (OV-Positioning) workflow and Skeleton-based 3D Pose (S3D-Pose) estimation workflow. Outside-in referred to the fact that an external camera was used to track the agents in the scene. The term \emph{outside-in} was borrowed from the augmented reality and virtual reality (AR/VR) community\footnote{AR/VR applications could either track the position of a head mounted display unit in a 3D world space using external sensors (outside-in) or using internal sensors (inside-out) equipped on the head-mounted display device.}.

This paper is the expanded version of our paper titled \emph{Visual-based Positioning and Pose Estimation} \cite{span20}. The contributions of this application paper are from: (i) the Outside-in Vision-based Positioning (OV-Positioning) workflow, and (ii) the Skeleton-based 3D Pose (S3D-Pose) estimation workflow. The rest of the materials in this paper are organized into the following sections; Section 2 describes related works; Section 3 describes the OV-Positioning, and the S3D-Pose workflows; Section 4 presents the evaluations, and Section 5 provides the conclusions and future direction.

\section{Related Works}
Fast R-CNN\cite{girshick15} and Faster R-CNN \cite{ren15} extend Region-based CNN (R-CNN) with an improved region proposal process that reduces the number of proposed regions while improving the relevancy of the proposals. Another augmentation over R-CNN is the introduction of the RoI pooling process. The RoI pooling layer takes feature maps from the previous CNN layer and transforms output feature maps to a fixed length vector then feed it to two fully connected layers which output bounding boxes and classes. A significant improvement of the Faster R-CNN over the Fast R-CNN is that the Region Proposal Network (RPN) in the Faster R-CNN is trained as part of the Faster R-CNN.

Mask R-CNN \cite{he18} further extends Faster R-CNN by adding a new segmentation masks output. The mask branch is a small Fully Convolutional Network (FCN) applied to each RoI, predicting a segmentation mask at a pixel level. RoIAlign is another important enhancement introduced in Mask R-CNN, and it shows a substantial performance improvement over the Faster R-CNN, which uses the RoI pooling process. The RoI pooling process commits to some information loss when the bounding box size is rounded up to integer values during the RoI pooling process. This drawback has been mitigated using the RoIAlign process, where information from four neighbour points is employed to compute the value of each sampling point without quantization.

\subsection{Human Activity Recognition}
Human Activity Recognition (HAR) is a popular open research problem. 
Recognizing human activities from a 2D image is a challenging problem since further detailed distinctions of objects in the same class must be identified, i.e., to identify various humans' actions from the same human class. Since image data does not explicitly encode activity information, relevant discriminative features must be extracted from image data first. This makes vision-based HAR (VHAR) a complex problem in comparison to formulating and solving HAR using other types of sensors, for example, using accelerometer data to track body or limbs movements, or using radio frequency tag (RFID) to track the place where activities are taking place. 
Although VHAR requires extra processing efforts to extract relevant discriminative features, it has many attractive points when its practicality is considered. Since humans recognize activities mainly from vision, the VHAR deployment often requires a minimum or no extra environmental preparation.

\subsubsection{Knowledge Representation}
VHAR problems are commonly formulated from the image recognition task perspective.
Representing an image as a feature vector derived from the whole image is a common option.  The original pixels information may also be preprocessed and transformed to enhance salient information: grayscale, binary silhouette for displaying the shape of an object, Histogram of Gradient (HOG), Scale Invariant Feature Transformation (SIFT); or be transformed to other representation schemes: space-time interest points (STIPs) \cite{laptev03}, sparse representations and, representation extracted from CNN feature maps \cite{mo16}.

A computer representation of an image can truthfully capture the intensity information of pixels but fails to capture the extra-meaning, which is the real semantic content of the image. On the contrary, humans can recognize information which is implicitly and explicitly encoded in an image, i.e., objects, their positions and interactions among objects. This fact is well appreciated, and contemporary image analysis research commonly focuses on low-level vision analysis. High-level activity analysis, such as VHAR is still developing where the recognition of activities from the skeleton is one of the major directions. Recent works in this direction capture activities using techniques such as concept hierarchical of actions \cite{kojima02}, an action graph model and a scene graph generation model \cite{xu17}.

Activities in a sport game can be described based on the agents' actions observed in each image, e.g., when the player jumps and returns the shuttlecock using a forceful stroke (in a badminton game). It seems intuitive that recognizing activities from skeleton pose should be more effective than modelling the recognition model from features extracted from the whole 2D image. 

\subsection{Human Skeleton Pose Estimation}
Motion Capture (MOCAP) is an example of a 3D pose estimation task. Optical based MOCAP systems determine 3D pose based on multiple cameras detecting markers attached to actors' key points, e.g., wrists, elbows, hands, knees, and head. 

In \cite{shotton13}, the authors demonstrated a real-time human pose estimation from a single depth image. The approach enabled accurate recognition of a limited set of activities which could be expressive enough for some game-based applications. Pose estimation from RGBD image has stirred up interest in performing 2D pose estimation using RGB image.

Object detection technique can be employed to detect human body key points such as head, shoulders, elbows, knees, without markers.  With advances of deep learning and computer vision techniques in the past decade, human pose estimation techniques have improved significantly. CNNs were employed to identify 2D key-points on the human body \cite{toshev14,tompson15,wei16,newell16} from unconstrained images. There are always many possible 3D interpretations of a 2D pose due to an extra dimension. Fortunately, physical constraints of human anatomy and intraframe information (in case of an image sequence) can be applied to filter out improbable 3D pose interpretations. In \cite{tome17,kudo18}, the authors demonstrated approaches to transform human pose in 2D space to 3D space using deep learning techniques.

\section{A Study of a Badminton Game Analyses}
Badminton is a popular indoor sport played on a 20 ft$\times$44 ft court. Two teams play against each other, and each team has one person (or two persons for a doubles match). Players use a racquet to hit a shuttlecock across the net. A team scores a point when the other side cannot return the shuttlecock, and it lands inside the court.

Traditional analysis of a badminton game is carried out manually by a human expert analyzing the movement of players, the opportunity to score and the errors made. Statistics of players' performance can reveal useful insights. Due to lacking  sophisticated video analysis techniques, this information is manually entered, and it is the bottleneck of the manual approach. Leverage on deep learning and computer vision techniques, players' movements and actions can be automatically extracted from video frames.

A more in-depth insight of the badminton game may be revealed by associating the position and action of players to their scoring performance. In this section, we are going to discuss the proposed pipeline of tasks that automate the position and pose analysis from an image sequence: OV-Positioning and S3D-Pose.

\begin{figure}[!ht]
\begin{center}\leavevmode
\epsfxsize=10cm
\epsfbox{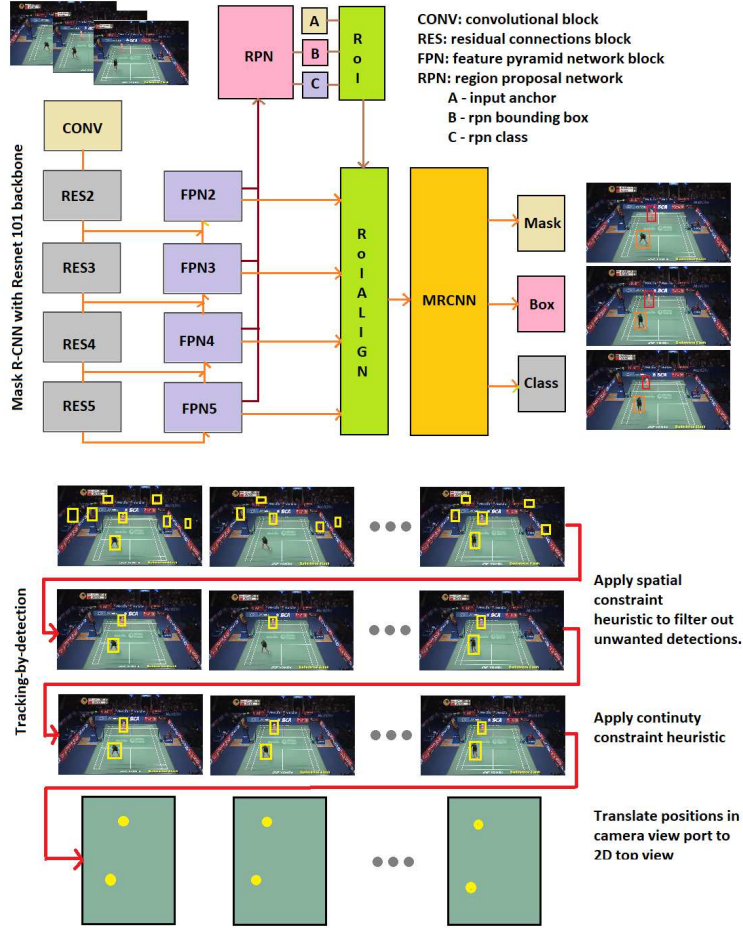}
\end{center}
\caption{Top pane: Image frames from a camera are processed by Mask R-CNN, players' positions and bounding boxes are determined. The three images on the right show example output of the detection from Mask R-CNN.
Bottom pane: Details the processing pipeline from 2D positions in camera view to a 2D top-view in the world space. Detection: Mask R-CNN detects multiple objects which could be noisy (top row).  Tracking-by-detection: various heuristics are applied to filter out undesired boxes (row 2 \& 3). At the final stage, players' positions are translated from the camera viewport into top-view in world space (bottom row).}
\label{ovbp}
\end{figure}

\subsection{Outside-in Vision-based Positioning Workflow (OV-Positioning)}
The OV-Positioning workflow consisted of the following tasks: object detection, tracking-by-detection, and translation to the positions in the world coordinates. Figure \ref{ovbp} shows a graphical summary of the OV-Positioning workflow. 

\subsubsection{Object Detection}
A sequence of image frames, each with 852 $\times$ 472 (width $\times$ height in pixels), passed through the object detection block. Here, Mask R-CNN with Resnet-101 backbone\footnote{matterport} was employed for human detection. ResNet is a convolutional neural network with skip connections. This seemingly simple modification improves learning and performance of the deep network significantly. Mask R-CNN is the current state-of-the-art semantic segmentation technique. It provides high-quality bounding box output of objects. This is achieved from the combination of Feature Pyramidal Network (FPN), Region Proposal Network (RPN) and RoIAlign which effectively exploit information in the feature maps at different scales and with precise pixel alignment (than older techniques in Faster R-CNN).

\subsubsection{Tracking-by-detection}
Tracking-by-detection operates on the assumption that object tracking can be simply derived from object detection, provided that an effective object detector can detect objects with good spatial accuracy and high resolution in time. In real life applications, detection errors should be expected and dealt with. In this badminton game analysis task, the Mask R-CNN returned many detected objects. Here, the following heuristics were implemented: (i) a spatial constraint was applied to filter out all detected objects which were not in the badminton court. This heuristic ensured that only two players were detected; (ii) a continuity in spatial space was applied across frames. This heuristic filtered out erroneously detected boxes that resulted in a big jump in spatial space, and finally (iii) there were only two detected players in each frame. This heuristic dealt with extra detected boxes (false positive) or missing detected boxes (false negative). The missing detection or extra detections could be dealt with based on the information of the boxes from other frames.

\subsubsection{Perspective Transformation Between Planes}

The position of a human in the scene on the floor (x,y) coordinate (i.e., a virtual top-view camera) can be transformed from the position obtained from the camera view. This process is known as homography transformation.
In the first stage, the players' positions ${\bf p}(x_c,y_c)$ were detected on the camera viewport. We apply the perspective transformation to transformed the camera viewport plane to its corresponding 2D top-view in the world coordinate ${\bf p}(x_w,y_w)$ using the homography matrix $H$; $P_w = HP_c$.

\begin{figure}[!ht]
\begin{center}\leavevmode
\epsfxsize=8cm
\epsfbox{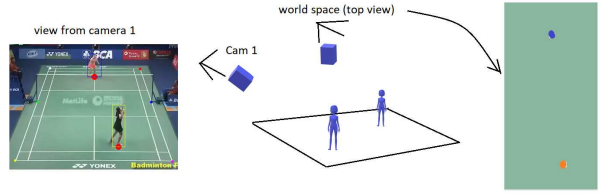}
\end{center}
\caption{With at least four known coordinates on the badminton court from the camera view and their corresponding coordinates in the world view, the homography matrix can be estimated and employed to compute the players' positions as if they were viewed from a virtual top-view camera.}
\label{ovbp3}
\end{figure}

The homography matrix was estimated based on at least four known coordinates of the badminton court in the camera view and the corresponding coordinates in the world view. Figure \ref{ovbp3} illustrates the transformation of the camera view to the world view.

\begin{figure}[!ht]
\begin{center}\leavevmode
\epsfxsize=9cm
\epsfbox{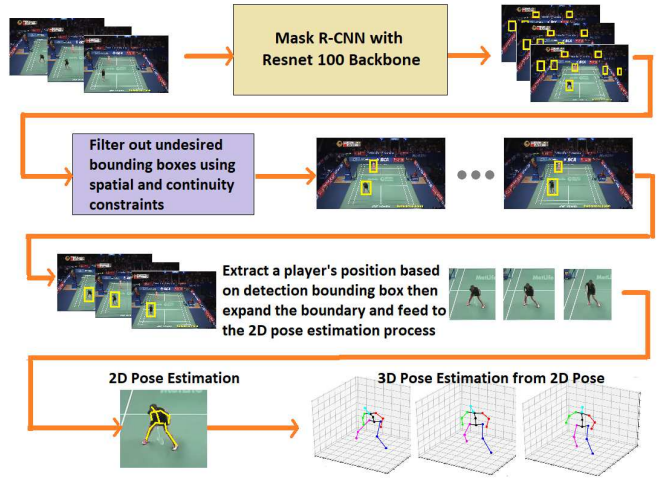}
\end{center}
\caption{The processing pipeline for 3D pose estimation: Mask R-CNN detects players location in the scene. The player bounding box is expanded, and key points in the 2D view are extracted. The 3D pose is estimated based on the 2D pose.}
\label{sbai}
\end{figure}

\subsection{Skeleton-based 3D Pose (S3D-Pose) estimation workflow}
A graphical summary of the S3D-Pose estimation workflow is shown in Figure \ref{sbai}.
The S3D-Pose estimation workflow consisted of the following tasks: object detection, tracking-by-detection, 2D pose estimation and 3D pose estimation from a 2D pose. It should be noted that the processing of Mask R-CNN, object detection, and tracking by detection processes were shared with the OV-Positioning discussed in the previous section. 

The image patch of a player was then extracted with an extra 30-pixels expansion in all four directions, i.e., up, down, left and right. This was to ensure that the player's full body was captured since the bounding box from the object detection process is commonly designed to fit the object. We leverage on the implementation from \cite{tome17}, \emph{Lifting from the Deep (LD)}, for the 2D and 3D pose estimation. LD fused probabilistic knowledge of 3D human pose with a multi-stage CNN architecture and used the knowledge of plausible 3D landmark locations to refine the search for better 2D locations. 

The 3D poses extracted from 2D poses might not be consistent. It was possible for 3D poses to go missing or be misinterpreted in some frames. We applies (i) linear interpolation to fill in the 3D poses from any given two keyframes; and (ii) inter-frames continuity constraint to fix the abnormal jerk of positions between consecutive frames.

\section{Evaluation of Position and Pose Estimation}
\begin{figure}[!ht]
\begin{center}\leavevmode
\epsfxsize=6cm
\epsfbox{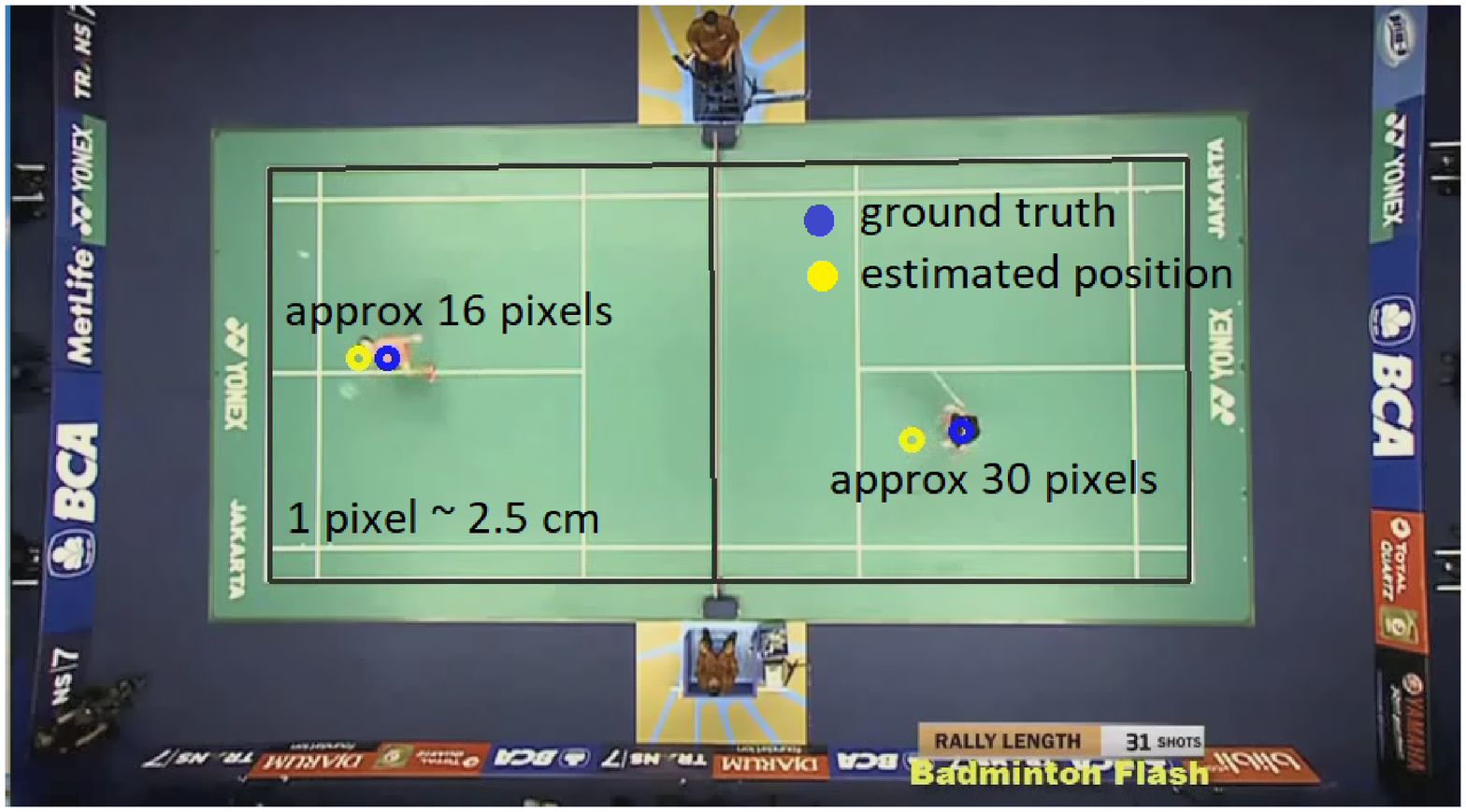}
\epsfxsize=5.5cm
\epsfbox{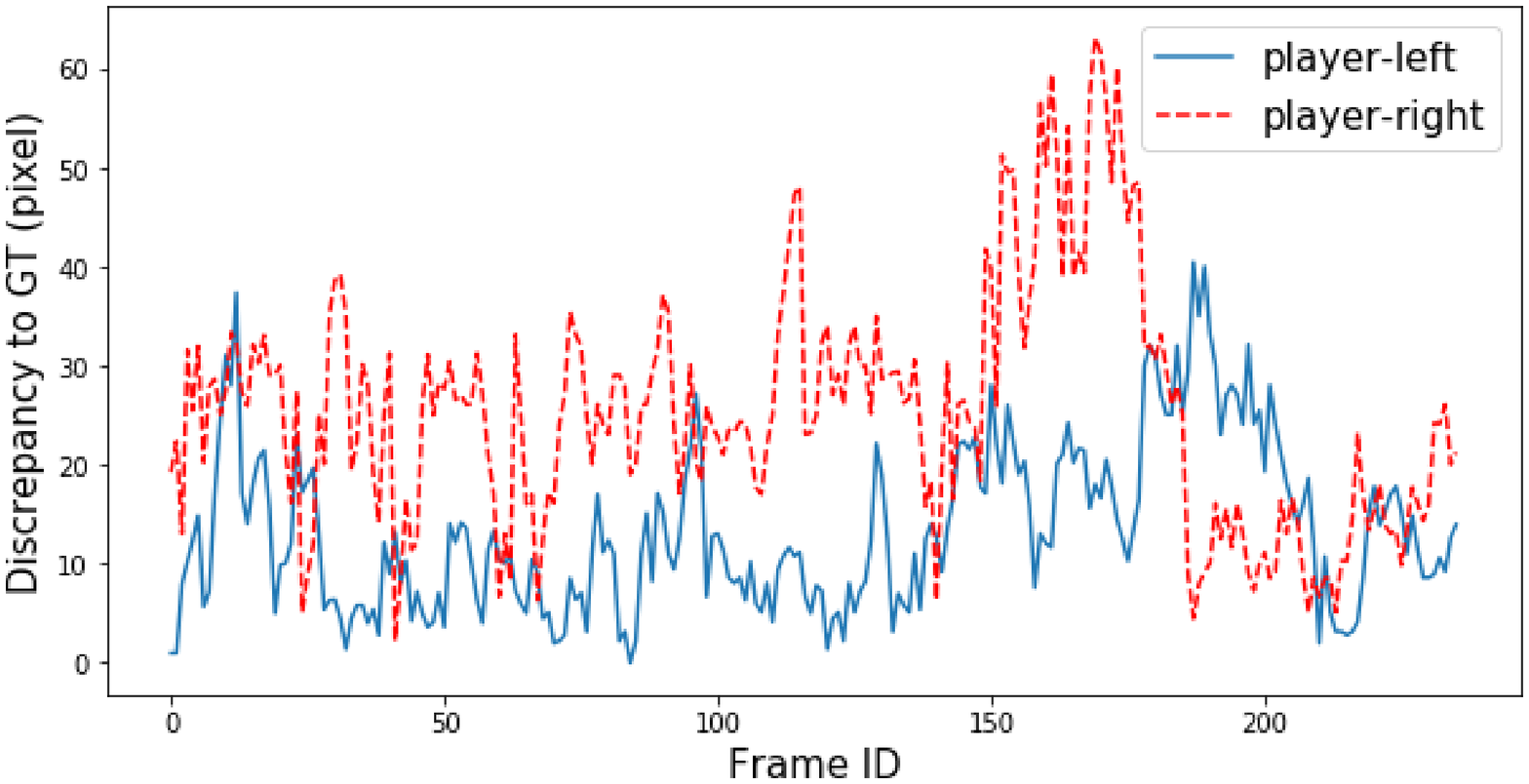}
\end{center}
\caption{The estimated positions obtained from the OV-Positioining process are compared with the ground truth positions. From the current setup, the discrepancy of one pixel is approximately 2.5 cm. The estimated positions of the player further from the camera are less accurate (the player on the right part of the frame).}
\label{eval1}
\end{figure}

\begin{figure}[!ht]
\begin{center}\leavevmode
\epsfxsize=5cm
\epsfbox{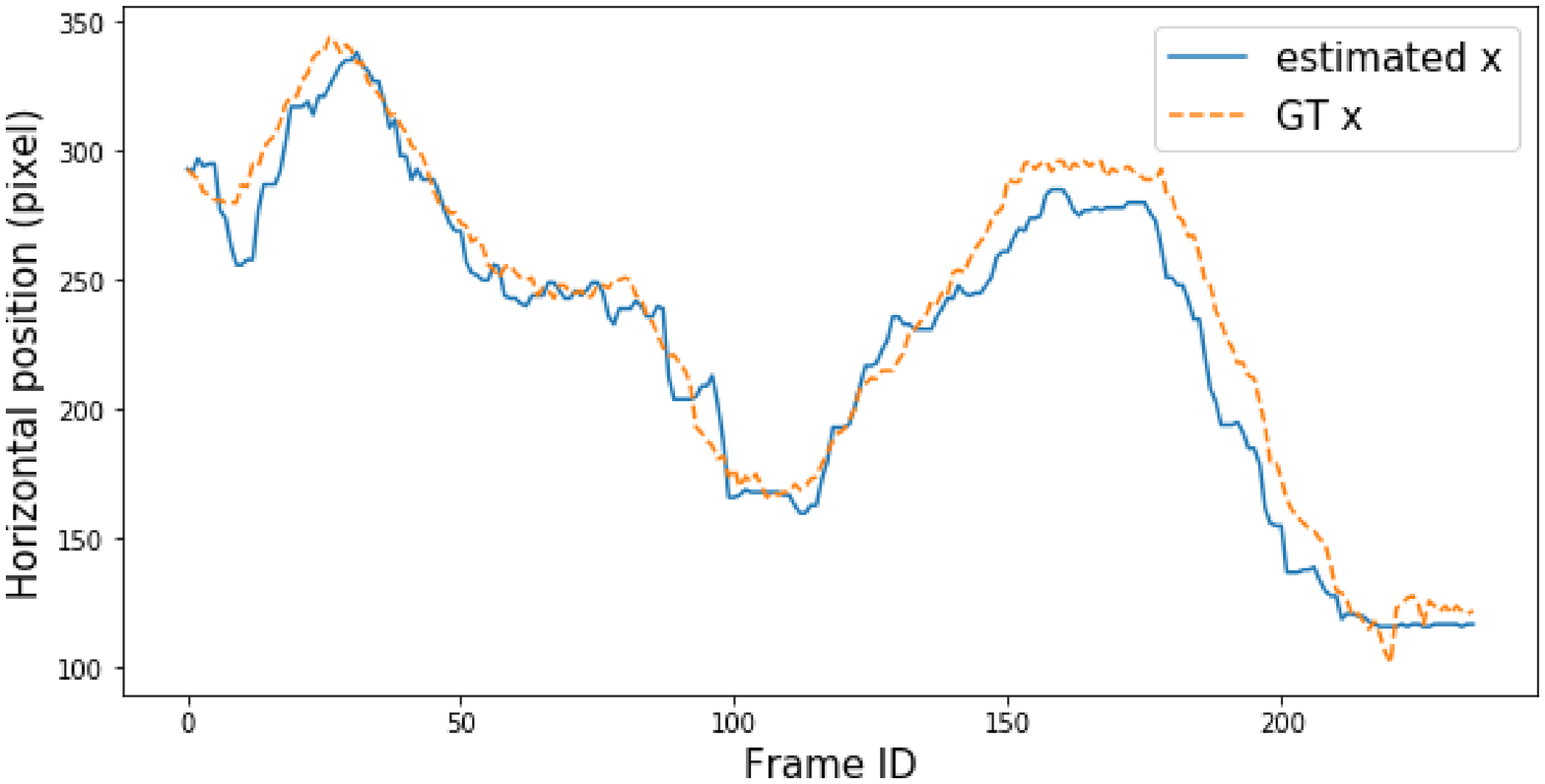}
\epsfxsize=5cm
\epsfbox{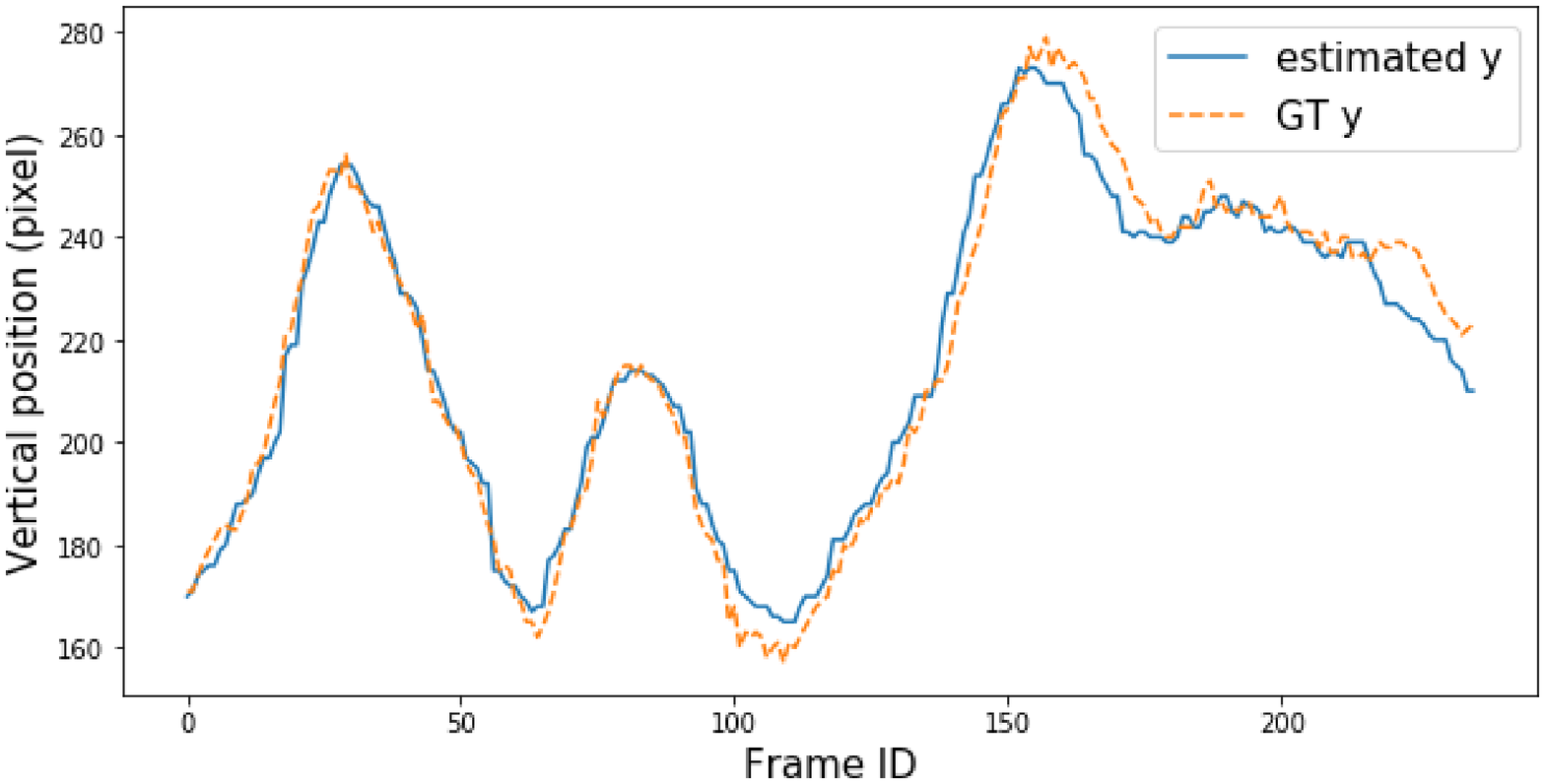}
\epsfxsize=5cm
\epsfbox{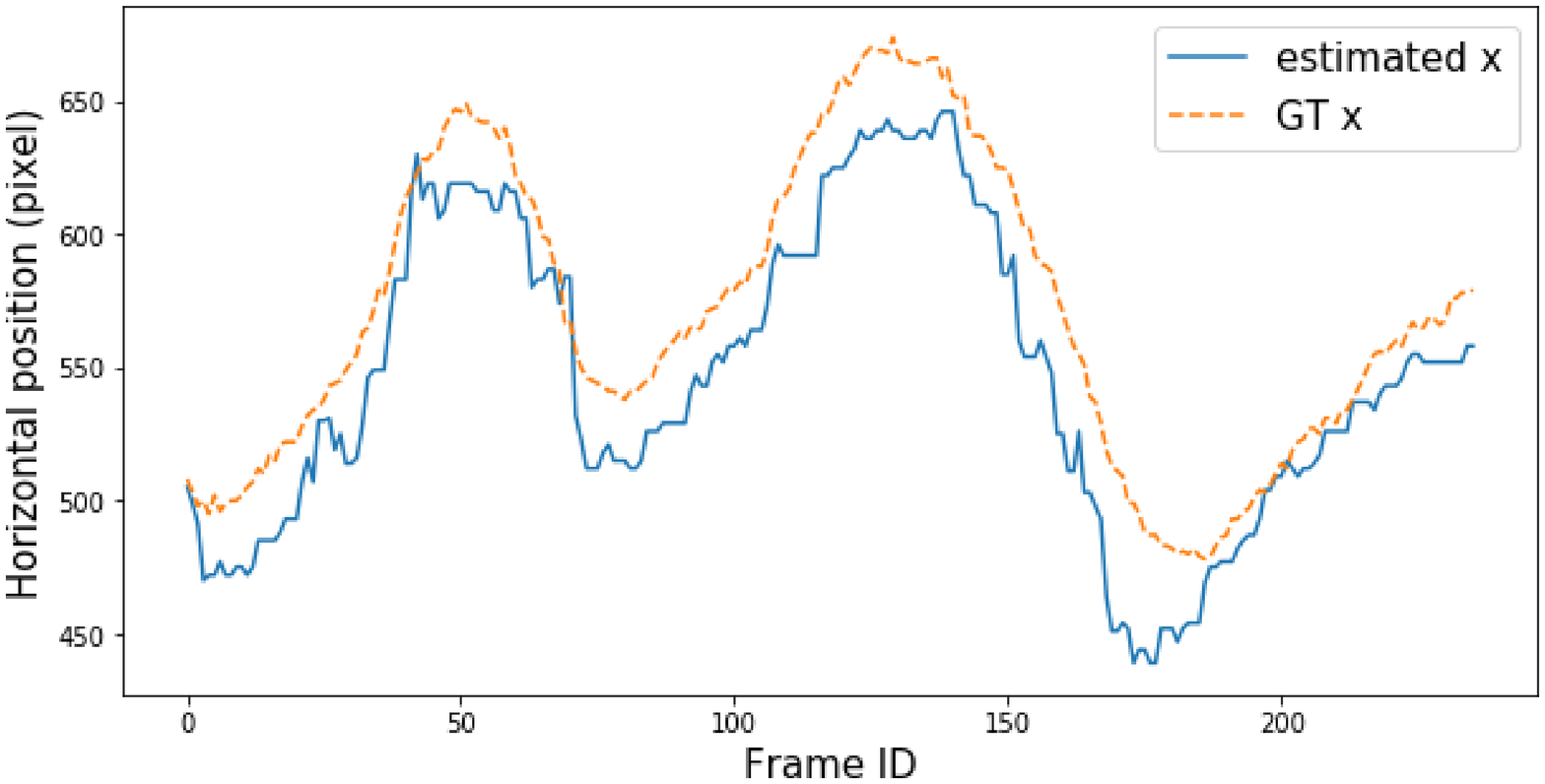}
\epsfxsize=5cm
\epsfbox{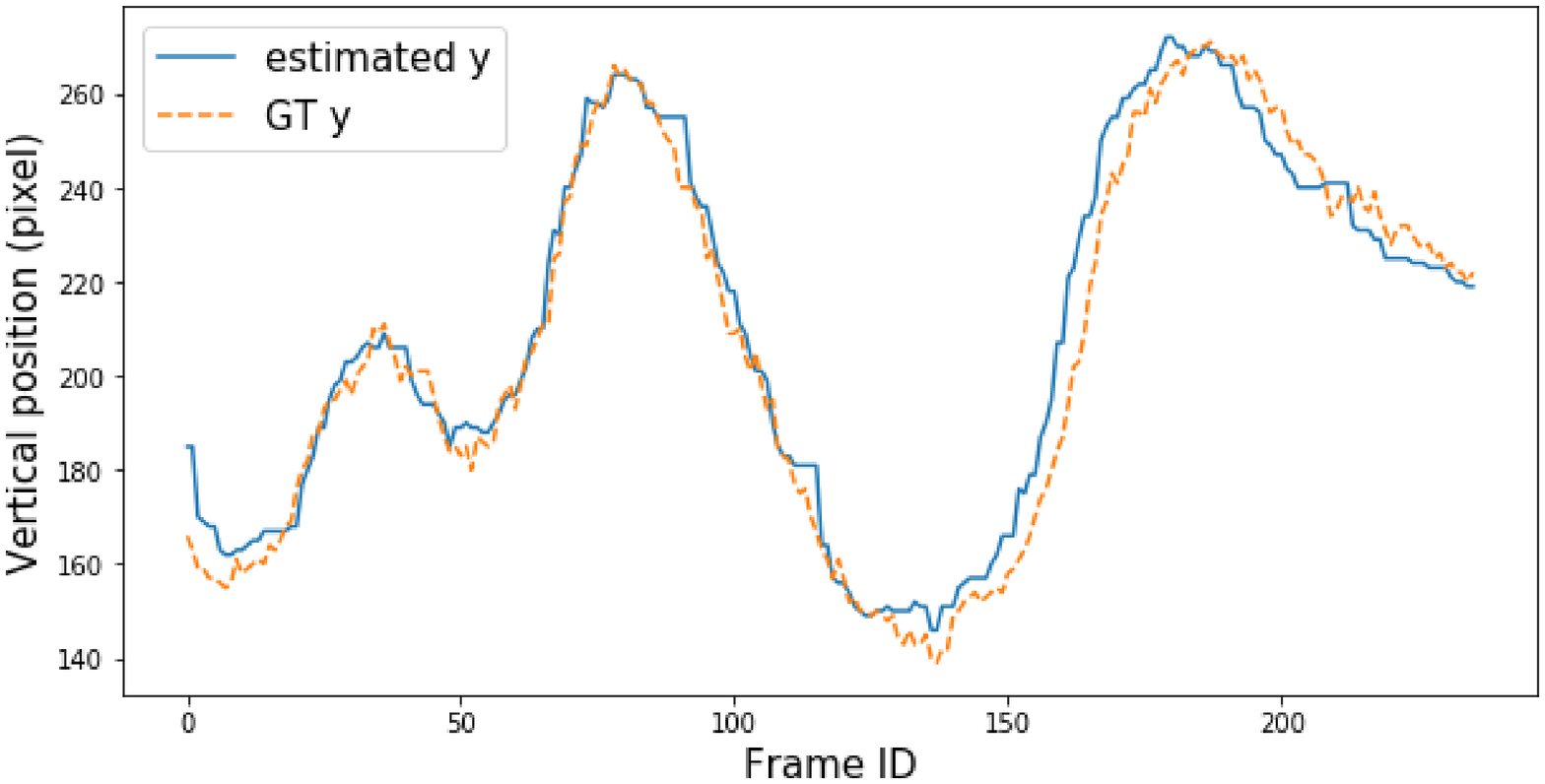}
\end{center}
\caption{Comparisons between estimated positions and ground truth positions, $x$ axis is frame ID, and $y$ axis is a discrepancy in pixel units. Top row: horizontal and vertical positions of the player on the left (near to the camera). Bottom row: horizontal and vertical positions of the player on the right (further from the camera).}
\label{eval2}
\end{figure}

We evaluated the output from the OV-Positioning workflow by comparing the projected position obtained from the homography transformation against the footage obtained from a top-view camera. Fortunately, some duels between the two players were also presented from a top-view camera. We manually marked the positions of the two players obtained from the top-view camera as a ground truth. The comparison between the projected positions and the ground truth positions are shown in Figures \ref{eval1} and \ref{eval2}. We discovered that the estimated positions of the players nearer to the camera had the average error of 13 pixels (equivalent to 32 cm. in our setup), and for the player further from the camera, the average error was  25 pixels (62 cm). Erroneous results were correlated with fast movements which could be easily mitigated with a higher frame sampling rate.

\begin{figure}[!ht]
\begin{center}\leavevmode
\epsfxsize=7cm
\epsfbox{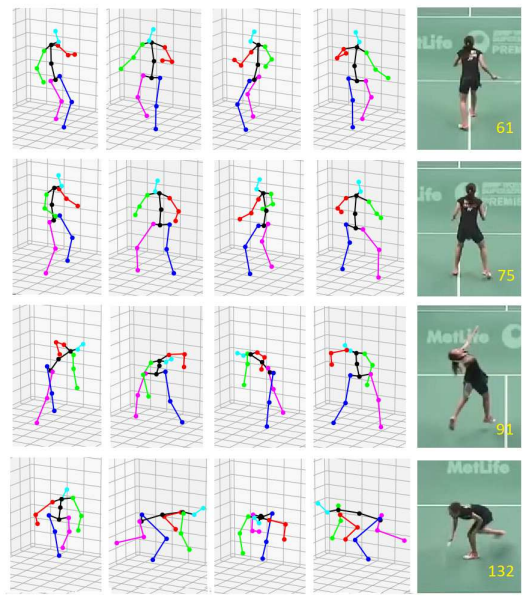}
\end{center}
\caption{The 3D skeleton lifted from the 2D skeleton which is interpreted from the 2D image on the rightmost column. Four skeletons are displayed at 30, 120, 210, and 300 azimuth degrees, respectively.}
\label{sbai1}
\end{figure}

\begin{figure}[!ht]
\begin{center}\leavevmode
\epsfxsize=9.5cm
\epsfbox{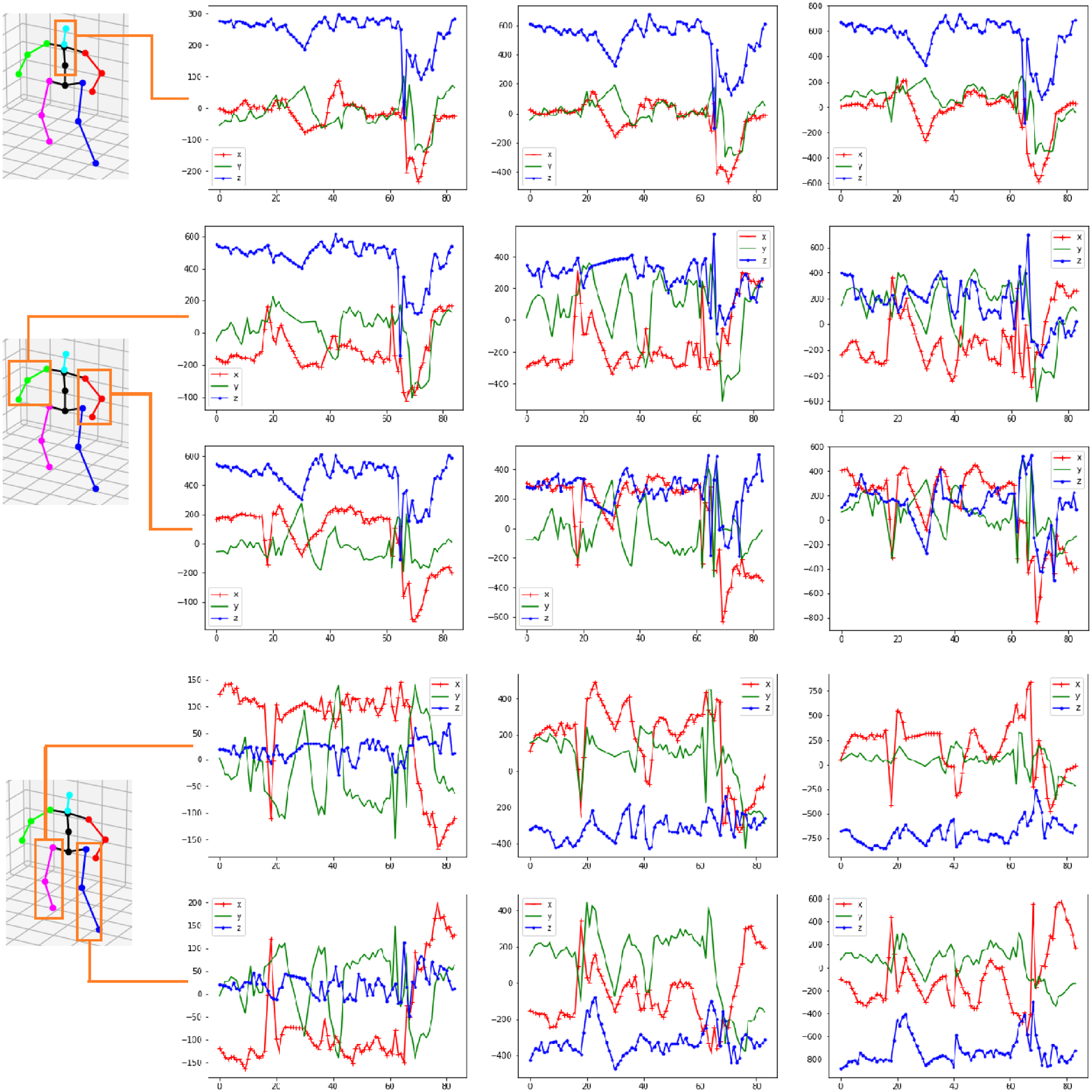}
\epsfxsize=10cm
\epsfbox{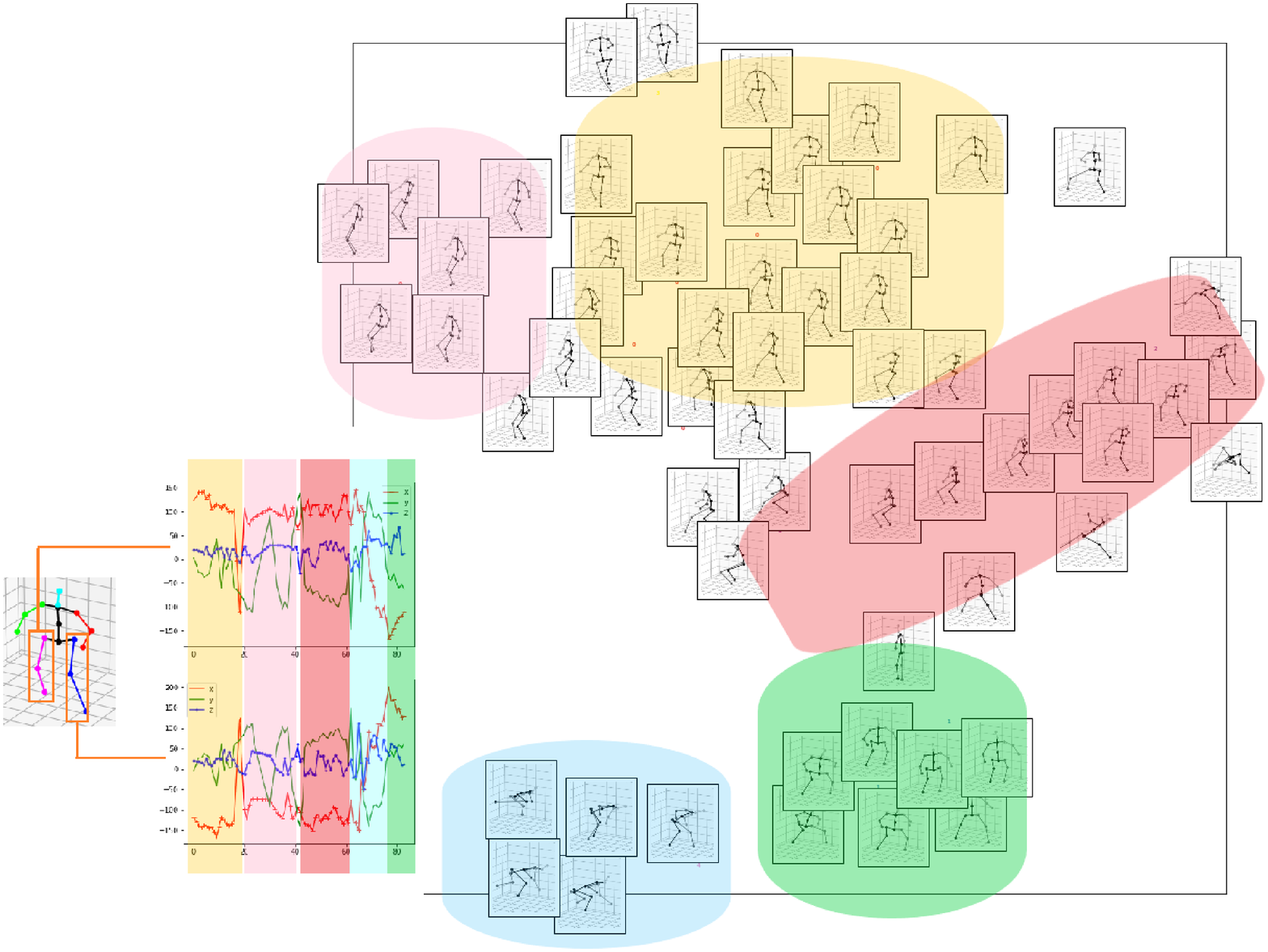}
\end{center}
\caption{Top pane: plots of $(x,y,z)$ coordinates from different key points from a short video segment. Bottom pane: the mapping of 51 features (from 17 key points) to 2D coordinate using t-SNE method, clusters of activities emerge.}
\label{sbai2}
\end{figure}

The results obtained from this study are very encouraging. We found that OV-Positioning could provide estimated positions with good temporal resolution and good spatial resolution. The drawbacks from the visual-based approach are from occlusion or when the object is far away from the  camera. These issues can be approached using a multi-camera system. 

Figure \ref{sbai1} shows the skeletons in 3D positions (at 30, 120, 210, and 300 azimuth degrees, respectively). Even though OV-positioning provided accurate positions of players such that appropriate human images were fed to the S3D-Pose for skeleton pose estimations, erroneous 3D poses were common and need to be cleaned up. We resorted to the \emph{inbetweening} technique to fill in-between frames between any two keyframes. Firstly, outlier frames were identified as erroneous frames using handcrafted constraints. The outlier frame was the frame with the improbable movement of keypoints from the previous frame to the next frame. After good keyframes and erroneous keyframes were identified, two suitable frames were taken as the two keyframes and the keypoints of the sandwiched outlier frame were then interpolated from the good keyframes. If there were many sandwiched frames, the interpolation was performed recursively by filling a frame in the middle point first then repeating the process with the newly added frame, now a new keyframe. 

Figure \ref{sbai2} (top pane) shows plots from the $(x, y, z)$ keypoints. These positions can be employed to train a classification model for activities recognition. We applied the \emph{t-distributed stochastic neighbor embedding} method to map 51 features (from 17 keypoints) in each frame to coordinates in a 2D space and plot corresponding activities, see Figure \ref{sbai2} (bottom pane). Clusters of various activities are clearly seen. One of the salient features of S3D-Pose is that the estimated 3D positions are invariant to different points of view. Hence, S3D-Pose is ideal for human activity analysis tasks that describe physical activities.

\section{Conclusion \& Future Directions}
Activities in a badminton game can be described based on the players' actions in each image, e.g., the player serves the shuttlecock. In this context, analyzing physical activities from the skeleton poses is more effective than confabulating an image caption using the similarity of image labels \cite{span19}.

We present two workflows: OV-Positioning and S3D-Pose for the position and pose analysis, respectively. The OV-Positioning estimates a position using visual information. It offers many salient features compared to other positioning approaches such as traditional indoor positioning systems (IPS) using WiFi or Bluetooth beacon technology. As discussed above, the OV-Positioning performs visual analyses to extract the agents' positions in the scene. Therefore, an extra WiFi system is not required to estimate their positions. The estimated positions from OV-Positioning have a better spatial resolution and with less latency than the estimation obtained from traditional wireless indoor position systems.
S3D-Pose estimates human pose in 3D. This offers many salient features since human pose in 3D are invariant to viewpoints. The approach is effective for physical activity identification task and could enable multi-disciplinary studies in various areas such as sport science and enhance broadcasting experience.

\begin{small}
\section*{Acknowledgments}
This publication is the output of the ASEAN IVO \texttt{http://www.} \texttt{nict.go.jp/en/asean\_ ivo/index.html} project titled  \emph{Event Analysis: Applications of computer vision and AI in smart tourism industry} and financially supported by NICT (\texttt{http://www.nict.go.jp}). We would also like to thank anonymous reviewers for their constructive comments and suggestions.
%

\end{small}
\end{document}